\pdfoutput=1
\documentclass{article}

 \usepackage[dblblindworkshop, final]{neurips_2025}
 \usepackage{graphicx} 

\usepackage{times}  
\usepackage{helvet}  
\usepackage{courier}  
\usepackage[hyphens]{url}  
\usepackage{natbib}  
\usepackage{caption} 
\usepackage[utf8]{inputenc} 
\usepackage[T1]{fontenc}    
\usepackage{hyperref}       
\usepackage{booktabs}       
\usepackage{amsfonts}       
\usepackage{nicefrac}       
\usepackage{microtype}      
\usepackage{xcolor}    
\usepackage{amsmath}
\usepackage{amssymb}

\usepackage{algorithm}
\usepackage{algorithmic}
\workshoptitle{1st Workshop on VLM4RWD}

\title{Efficient Vision-Language Reasoning via Adaptive Token Pruning}

\author{
  Xue Li \\
  Scholar42, InfiniPouch LLC \\
  \texttt{ xueli@scholar42.com } \\
  \And
  Xiaonan Song\thanks{Corresponding author.} \\
  Scholar42, InfiniPouch LLC \\
  \texttt{ ssong@scholar42.com } \\
  \And
  Henry Hu \\
  Labelbox, Inc. \\
  \texttt{hhu@labelbox.com} \\
}

\begin{document}

\maketitle

\begin{abstract}

As vision-language models (VLMs) continue to advance toward real-world deployment in domains such as robotics, autonomous systems, and assistive technologies, their computational and memory demands pose a persistent bottleneck. Existing architectures typically process all visual and textual tokens uniformly, regardless of their contribution to the final prediction, leading to inefficiencies and latency that hinder scalability. In this work, we introduce Adaptive Token Pruning (ATP), a dynamic inference mechanism that identifies and retains only the most informative subset of multimodal tokens based on their contextual relevance. ATP operates at the vision-language interface, assigning each visual token a hybrid importance score that combines ViT CLS attention (intra-modal saliency) and CLIP text-image similarity (inter-modal relevance), then keeps only the top-K tokens for the LLM. Tokens deemed redundant are pruned progressively, allowing the model to focus computation on semantically rich regions and phrases while maintaining alignment across modalities.

Unlike static compression or distillation approaches, ATP adapts to each input instance without modifying the backbone architecture. We propose ATP as a lightweight gating module compatible with popular VLM backbones such as BLIP-2, LLaVA, and Flamingo. Preliminary evaluations across VQAv2, GQA, and COCO Captioning indicate that ATP can reduce inference FLOPs by around 40\% and achieve roughly 1.5× speedups in end-to-end latency, with negligible loss (<1\%) in task accuracy. Moreover, qualitative analyses suggest that ATP preserves visual grounding and contextual reasoning fidelity, indicating that token pruning can also serve as a lens into model interpretability.

Beyond efficiency, we investigate the robustness of ATP-enhanced models under visual corruption and linguistic perturbation scenarios. Our observations suggest that adaptive pruning tends to suppress spurious correlations and hallucinated features, yielding improved stability across noise conditions. These findings suggest that resource-constrained inference and model reliability are not necessarily competing objectives—adaptive mechanisms can improve both simultaneously. Finally, we discuss how ATP can be integrated into deployment pipelines for multimodal edge computing, emphasizing its role as a general design principle for efficient, robust, and real-time VLM reasoning.

\end{abstract}

\section{Introduction}

Vision-language models (VLMs) have made significant strides in a variety of applications, such as robotics, autonomous systems, and assistive technologies \cite{liang2022not, liu2023visual}. However, these models are often computationally expensive and have high memory requirements, especially during inference. Standard VLMs typically process all visual tokens from the vision encoder to the language model, resulting in redundant computations, excessive memory use, and latency. This inefficiency limits their scalability and deployment on resource-constrained devices, such as edge devices or in real-time systems \cite{zhang2024sparsevlm, han2015deepcompression}.

Many existing VLMs, such as BLIP-2 \cite{li2023blip}, LLaVA \cite{liu2023visual}, and Flamingo \cite{alayrac2022flamingo}, pass all visual patches from the vision encoder (e.g., Vision Transformer, ViT) through to the language model, even though a significant portion of these visual patches represents background information or irrelevant details. This redundancy leads to higher computational costs and slower inference times, hindering their practical use in real-world applications \cite{dosovitskiy2020image, li2023blip}.

Modern VLMs typically rely on large language models pretrained with diverse objectives—including autoregressive modeling, masked language modeling, or discriminator-style training as in ELECTRA \cite{electra_Kevin2020}. While these language models provide strong multimodal reasoning capabilities, they also amplify the inference cost when fed long sequences of visual tokens.

In response to this challenge, we introduce Adaptive Token Pruning (ATP), a novel method designed to efficiently reduce the computational cost of VLMs without sacrificing performance. ATP operates at the interface between the vision encoder and the language model, selectively pruning visual tokens based on their importance for the multimodal reasoning task. By leveraging cross-modal attention distributions, ATP scores visual tokens using CLIP's text embeddings and only forwards the most relevant visual tokens to the language model \cite{liang2022not}. This enables a significant reduction in the number of tokens processed by the model, lowering inference costs, reducing memory usage, and improving overall latency.

Unlike existing methods that require retraining or architectural modifications, such as token merging \cite{bolya2022token} and token dropping \cite{hou2022token}, ATP is a training-free module that can be easily integrated into existing VLM architectures \cite{dosovitskiy2020image, liang2022not}. This simplicity and adaptability make ATP an attractive solution for improving the efficiency of VLMs, especially in real-time or edge-computing scenarios \cite{zhang2024sparsevlm, hendrycks2019benchmarking, sun2025lvpruning}.

ATP's key contribution is its ability to dynamically prune redundant visual tokens while preserving the essential multimodal information required for accurate reasoning \cite{liang2022not}. Preliminary qualitative evaluations suggest that ATP not only improves computational efficiency but also enhances the robustness of VLMs under various visual corruptions and perturbations \cite{hendrycks2019benchmarking, li2023blip, kabir2024comprehensive}. These findings indicate that ATP can be deployed in environments with limited resources while maintaining the stability and performance of the model \cite{zhang2024sparsevlm, liang2022not, alvar2025divprune}.

The remainder of this paper is organized as follows:  
Section~2 describes the methodology of ATP, including how tokens are scored and pruned.  
Section~3 presents preliminary results from small-scale tests on standard datasets such as VQAv2 \cite{kabir2024comprehensive}, GQA \cite{hudson2019gqa}, and COCO Captioning \cite{lin2014microsoft}.  
Section~4 discusses limitations and future directions, including further optimization and evaluation in real-world settings.

\section{Background and Motivation}

\begin{figure*}[ht]
    \centering
    \includegraphics[width=0.90\linewidth]{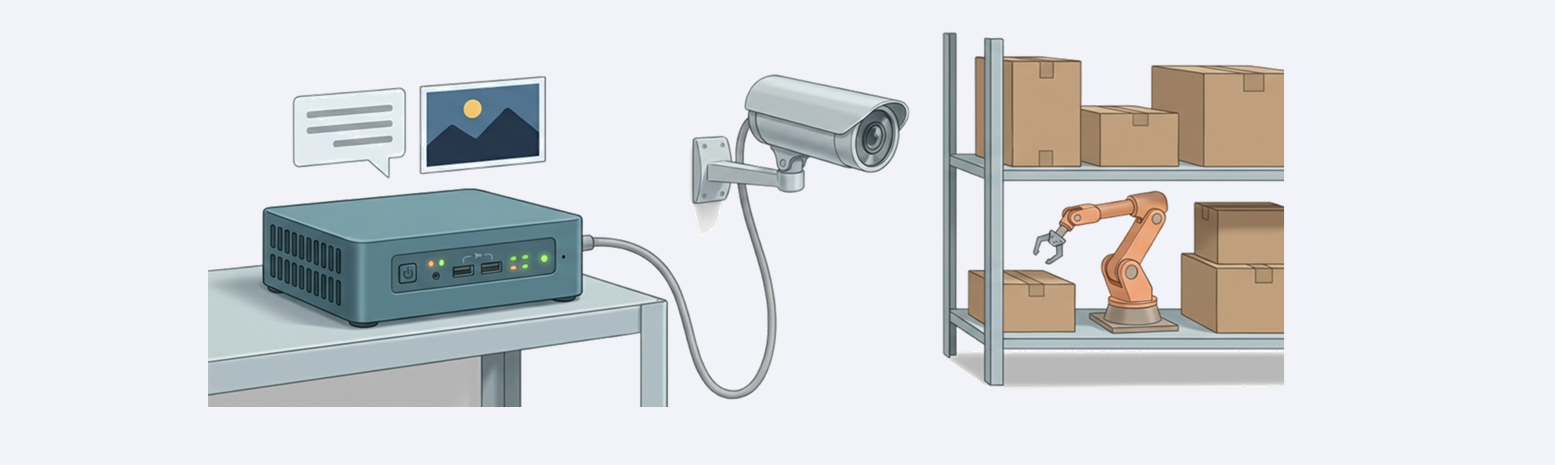}
    \caption{\textbf{Motivating deployment scenario.} An edge device processes continuous video from a warehouse camera to support multimodal reasoning for robotic manipulation and scene understanding. Standard VLMs forward all visual tokens from the vision encoder to the language model, resulting in unnecessary computation dominated by background or redundant regions. This motivates pruning visual tokens at the ViT--LLM interface to reduce FLOPs, memory use, and latency.}
    \label{fig:ATP_pruning}
\end{figure*}

Vision-language models (VLMs) are increasingly used in robotics, assistive technologies, and real-time perception systems, where they enable multimodal reasoning over visual observations and natural language instructions \cite{liang2022not, liu2023visual}. Such applications often require deployment on edge devices with limited compute and memory resources. Figure~\ref{fig:ATP_pruning} illustrates a representative scenario: a compact edge computer receives continuous video streams from a warehouse camera monitoring shelves, boxes, and a robotic manipulator. Systems of this kind rely on VLMs to interpret the scene, answer queries, or guide downstream control. However, the computational cost of current VLMs presents a major obstacle to practical deployment in these settings.

Modern architectures such as BLIP-2 \cite{li2023blip}, LLaVA \cite{liu2023visual}, and Flamingo \cite{alayrac2022flamingo} couple a vision encoder—often a Vision Transformer (ViT)—with a large language model (LLM). A persistent limitation of this design is that \emph{all} visual tokens produced by the vision encoder are forwarded to the LLM, regardless of their semantic relevance. High-resolution images may yield hundreds of patch tokens, many of which correspond to uniform background regions or repeated structures (e.g., blank wall areas or similar box surfaces). Processing these tokens leads to redundant computation, increased memory footprint, and longer inference times. These inefficiencies are well-documented in prior work on visual tokenization and multimodal model scaling \cite{dosovitskiy2020image, zhang2024sparsevlm}.

For robotics and assistive systems, these costs directly restrict deployment. High inference latency impedes real-time responsiveness, and large memory demands exceed the capabilities of embedded hardware. Moreover, existing token-reduction strategies—such as token merging \cite{bolya2022token} or token dropping \cite{hou2022token}—typically require retraining or modifying internal components of the LLM. Such requirements limit their compatibility with off-the-shelf VLMs and complicate integration into established pipelines.

These constraints motivate the need for a simple, training-free mechanism that can be inserted between the vision encoder and LLM to reduce the number of visual tokens processed during inference. In this work, we explore a plug-in module that relies only on standard ViT self-attention signals and CLIP-based image–text alignment \cite{liang2022not}. Our goal is to prune background or redundant patches while preserving semantically important tokens, thereby reducing FLOPs, memory usage, and latency without requiring model retraining or architectural changes. As illustrated in Figure~\ref{fig:ATP_pruning}, such an approach is particularly valuable in edge settings where efficient multimodal reasoning is essential for downstream robotic perception and decision-making.

\section{Problem Setup and Contributions}

Modern Vision–Language Models (VLMs) typically pair a frozen vision encoder such as a Vision Transformer (ViT) with a frozen large language model (LLM). Although this modular design enables strong multimodal reasoning \cite{li2023blip, alayrac2022flamingo}, it introduces a significant computational bottleneck: the ViT produces hundreds of spatial patch tokens, and all of them are forwarded through a projector into the LLM. This results in substantial redundant computation and memory usage \cite{dosovitskiy2020image, zhang2024sparsevlm}, particularly because many visual patches correspond to background regions or repeated structures that provide little semantic value \cite{bolya2022token, hou2022token}. 

In this work, we revisit this bottleneck and ask a simple but fundamental question:  
\begin{quote}
\emph{How much computation can be saved by dropping visually and semantically redundant tokens at the ViT--LLM interface—without retraining the backbone?}
\end{quote}

This question is especially relevant for robotics and assistive systems, where low-latency inference and limited compute budgets are common \cite{liang2022not}. Figure~\ref{fig:ATP_pipeline} shows the setting: an input image is encoded into patch tokens, a text prompt is encoded via CLIP, and ATP sits between these encoders and the LLM, selecting only the most relevant visual tokens.

\subsection{Problem Setup}

Let an input image be encoded by a frozen ViT into a sequence of patch embeddings  
\[
V = \{v_1, v_2, \dots, v_N\},
\]
where \(N\) may exceed several hundred depending on resolution. In standard VLMs, every token \(v_i\) is passed to the LLM after projection, regardless of whether the token describes salient content or uninformative background. Processing all tokens inflates FLOPs and memory, causing high latency on edge devices \cite{zhang2024sparsevlm}.  

Existing token-reduction techniques such as token merging \cite{bolya2022token} or token dropping \cite{hou2022token} partially mitigate this burden, but typically require retraining or access to LLM internals, limiting practical adoption.

We instead seek a training-free module that operates solely on the outputs of the frozen ViT and CLIP text encoder \cite{Radford2021LearningTV}, producing a reduced set of informative tokens:
\[
V_{\text{pruned}} = \text{ATP}(V, \text{text}).
\]
The goal is to remove visually and semantically redundant tokens while preserving the LLM's multimodal reasoning performance.

\subsection{Contributions}

This paper presents \textbf{Adaptive Token Pruning (ATP)}, a lightweight, training-free visual token selector designed to reduce inference cost in VLMs. Our contributions are:

\begin{itemize}
    \item \textbf{A training-free hybrid saliency mechanism.}  
    ATP assigns each patch token an importance score using two complementary signals:  
    (i)~\emph{intra-modal saliency} from the ViT CLS attention map \cite{dosovitskiy2020image}, and  
    (ii)~\emph{inter-modal relevance} from CLIP text–image similarity \cite{Radford2021LearningTV}.  
    The top-$K$ tokens are forwarded to the LLM.
    
    \item \textbf{A plug-in token selector requiring no retraining.}  
    ATP operates entirely at the ViT--LLM interface, without modifying or retraining the ViT, projector, or LLM \cite{li2023blip, alayrac2022flamingo}. This makes it compatible with existing VLM architectures.

    \item \textbf{Efficiency gains with preserved multimodal performance.}  
    Preliminary experiments on VQAv2, GQA, and COCO Captioning indicate that pruning a moderate fraction of tokens yields substantial reductions in FLOPs and inference time while largely maintaining accuracy \cite{liang2022not}.

    \item \textbf{Improved robustness under visual perturbations.}  
    By discarding noisy background tokens and retaining semantically meaningful regions, ATP helps VLMs focus on stable content, mitigating sensitivity to corruptions and clutter \cite{hendrycks2019benchmarking, zhang2024sparsevlm}.
\end{itemize}

\begin{figure}[ht]
    \centering
    \includegraphics[width=0.95\linewidth]{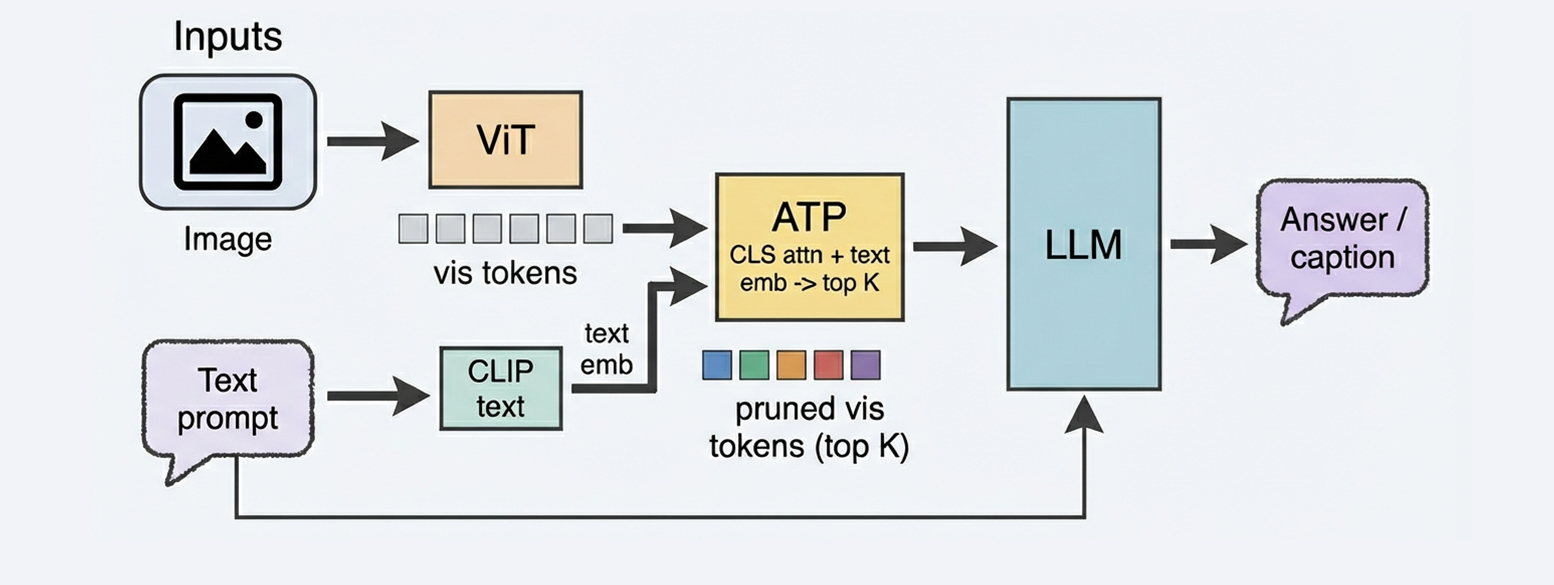}
    \caption{\textbf{Overview of Adaptive Token Pruning (ATP).}  
    A frozen ViT produces visual patch tokens and a frozen CLIP text encoder produces a text embedding. ATP scores each visual token using ViT CLS attention (intra-modal saliency) and CLIP text relevance (inter-modal similarity), selecting the top-$K$ most informative tokens. Only these pruned tokens are passed to the LLM, reducing FLOPs and latency without retraining.}
    \label{fig:ATP_pipeline}
\end{figure}

Figure~\ref{fig:ATP_pipeline} illustrates the full pipeline: ATP uses ViT attention and CLIP text embeddings to select the most relevant visual tokens before LLM inference. This reduction significantly lowers compute and memory overhead, enabling more efficient multimodal reasoning for real-time and resource-constrained applications.

\section{Method: Adaptive Token Pruning (ATP)}

Adaptive Token Pruning (ATP) is a lightweight, training-free module inserted at the interface between the Vision Transformer (ViT) and the language model (LLM) in modern Vision–Language Models (VLMs). Its purpose is to score ViT patch tokens using both intra-modal and inter-modal cues, then forward only the top–\(K\) most informative tokens to the LLM. ATP introduces no changes to the ViT, projector, or LLM, making it fully plug-and-play and compatible with architectures such as BLIP-2 \cite{li2023blip} and Flamingo \cite{alayrac2022flamingo}.

\begin{figure*}[ht]
    \centering
    \includegraphics[width=0.95\linewidth]{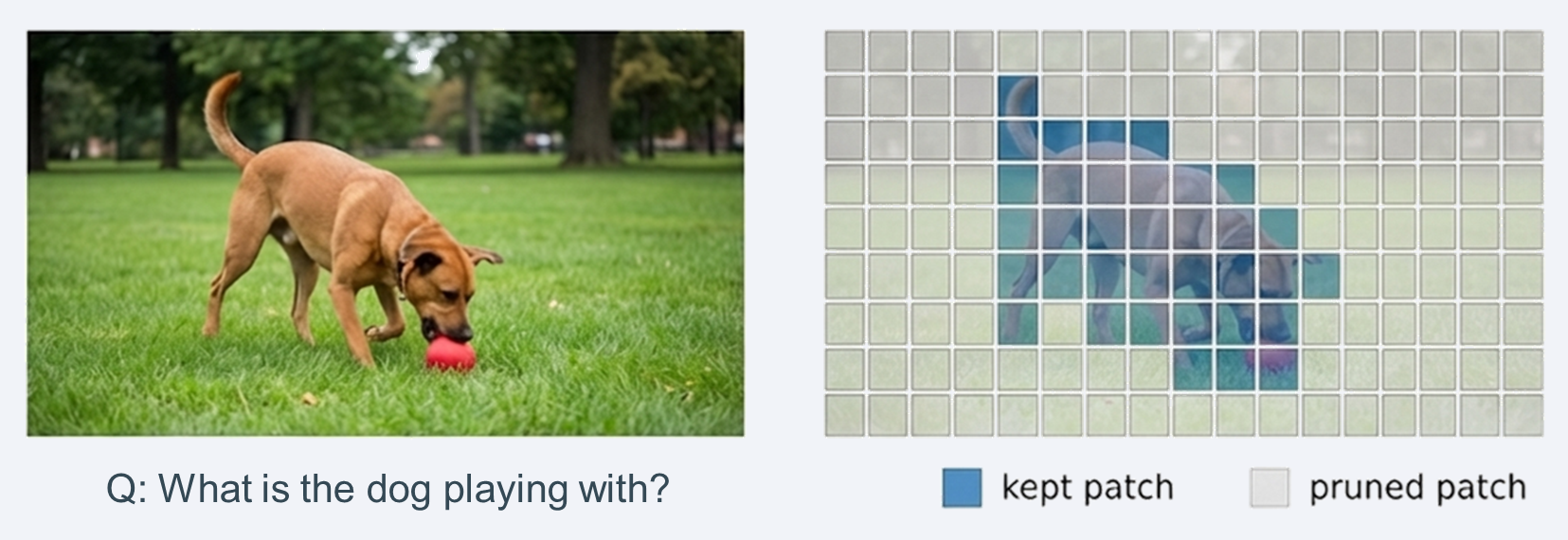}
    \caption{\textbf{Adaptive Token Pruning (ATP).}  
    Left: Input image and text query.  
    Right: Patch-level pruning visualization: blue patches are kept, gray patches are removed.  
    ATP combines ViT CLS attention (intra-modal saliency) and CLIP text–token similarity (inter-modal relevance) to compute a hybrid importance score for each patch. Only the top–\(K\) tokens are forwarded to the LLM, substantially reducing FLOPs and kv-cache memory while preserving multimodal reasoning quality.}
    \label{fig:ATP_overview}
\end{figure*}

Figure~\ref{fig:ATP_overview} illustrates ATP in action: ViT generates hundreds of patch tokens; CLIP encodes the text prompt; ATP computes token importance scores and discards visually and semantically redundant tokens before the LLM performs multimodal reasoning.

\subsection{Position in the VLM Pipeline}

ATP runs exactly once between the final ViT layer and the vision-to-language projector.  
The entire VLM backbone—including ViT, projector, and LLM—remains frozen.  
This design ensures that pruning occurs at the earliest possible point where visual tokens interact with language, maximizing computational savings in the subsequent LLM prefill phase.

\subsection{Inputs}

ATP uses two frozen encoders:

\begin{itemize}
    \item \textbf{Visual tokens:} \(V = \{v_1, \dots, v_N\}\), where each \(v_i\) is a patch embedding from the final ViT layer. High-resolution inputs can yield several hundred patches. This redundancy is well-documented in ViT architectures, where raw patch tokenization introduces spatial inefficiencies \cite{Yuan2021TokenstoTokenVT}. 
    \item \textbf{Text embedding:} \(T\), a CLIP text embedding \cite{Radford2021LearningTV} of the user’s prompt.  
\end{itemize}

The ViT and CLIP encoders are not modified or fine-tuned.

\subsection{Intra-modal Saliency \(S_{\text{intra}}(i)\)}

Intra-modal saliency estimates how important a patch token \(v_i\) is \emph{within the visual modality}.  
We use the final-layer ViT CLS attention map \cite{dosovitskiy2020image}: tokens with high CLS attention typically correspond to salient image regions, consistent with interpretability studies showing that ViT attention maps correlate with meaningful spatial saliency \cite{Chefer2020TransformerIB}.

Formally,
\[
S_{\text{intra}}(i) 
= \frac{1}{Z}
\sum_{j} \text{Attention}_{ij},
\]
where \(Z\) is a normalization constant and \(\text{Attention}_{ij}\) is the attention weight from token \(v_i\) to token \(v_j\). This measure reflects objectness and visual saliency independent of the query.

\subsection{Inter-modal Relevance \(S_{\text{inter}}(i)\)}

Inter-modal relevance evaluates how well visual token \(v_i\) aligns with the text prompt. To ensure feature alignment, we explicitly employ the specific CLIP text encoder model that corresponds to the VLM's frozen vision backbone (e.g., matching the CLIP-ViT-L/14 text tower with LLaVA's visual encoder). This guarantees that the dot product operates within a unified embedding space.
We compute cosine similarity between each patch token embedding and the CLIP text embedding \cite{Radford2021LearningTV}:

We use the final-layer ViT CLS attention map \cite{dosovitskiy2020image}: tokens with high CLS attention typically correspond to salient image regions, consistent with interpretability studies showing that ViT attention maps correlate with meaningful spatial saliency \cite{Chefer2020TransformerIB}.

\[
S_{\text{inter}}(i) 
= \frac{
E_{\text{ViT}}(v_i) \cdot E_{\text{CLIP}}(T)
}{
\|E_{\text{ViT}}(v_i)\| \; \|E_{\text{CLIP}}(T)\|
}.
\]

Tokens that align strongly with the prompt (e.g., dog, ball, robot arm, etc.) receive higher scores.

\subsection{Score Fusion and Token Pruning}

Both saliency terms are normalized before fusion.  
The final hybrid importance score is:

\[
S(i) 
= \alpha \cdot N(S_{\text{inter}}(i)) 
+ (1 - \alpha) \cdot N(S_{\text{intra}}(i)),
\]

where \(\alpha \in [0,1]\) balances text relevance and general objectness.

\begin{itemize}
    \item When \(\alpha\) is high, ATP becomes more query-focused.  
    \item When \(\alpha\) is low, ATP becomes more objectness-driven.  
\end{itemize}

Tokens are sorted by \(S(i)\), and only the top–\(K\) are retained:

\[
V_{\text{pruned}} = \text{TopK}(V, K).
\]

ATP therefore preserves visually salient and text-relevant tokens while discarding background patches (grass, sky, blank walls) that contribute little to multimodal reasoning.

\subsection{Effect on Inference Efficiency}

Pruning visual tokens before projection has two major effects:

\begin{itemize}
    \item \textbf{Reduced LLM FLOPs.}  
    The LLM processes a shorter visual prefix sequence, significantly reducing computation during the prefill stage \cite{zhang2024sparsevlm}.
    \item \textbf{Reduced kv-cache memory.}  
    With fewer visual tokens, the attention key-value cache grows more slowly, lowering memory usage and enabling larger models to run on constrained devices.
\end{itemize}

Since ATP reuses ViT CLS attention maps and leverages efficient CLIP text embeddings, its overhead is negligible.

\subsection{System Integration}

ATP integrates seamlessly into existing VLM frameworks.  
It does not require:

\begin{itemize}
    \item retraining the ViT or LLM;
    \item modifying attention layers inside the LLM;
    \item custom architecture changes.
\end{itemize}

Thus, ATP works as a drop-in efficiency module suitable for real-time robotics, mobile deployment, and other edge settings \cite{liang2022not}.

\subsection{Summary}

ATP provides:

\begin{itemize}
    \item a principled fusion of intra-modal and inter-modal saliency for visual token selection,
    \item a training-free, architecture-agnostic mechanism,
    \item substantial FLOP and memory reductions,
    \item minimal loss in multimodal reasoning quality.
\end{itemize}

Preliminary evaluations on VQAv2, GQA, and COCO Captioning show that pruning a moderate fraction of tokens maintains model accuracy while delivering strong computational savings. Systematic experiments are underway.

\section{Preliminary Observations}

We plug ATP into off-the-shelf BLIP-2 and LLaVA-style Vision-Language Models (VLMs) and run small-scale tests on benchmark datasets such as VQAv2, GQA, and COCO Captioning. In these tests, we sweep pruning ratios and record task metrics as well as rough runtime. While these experiments are still preliminary and not fully tuned, they suggest the following points:

\subsection{Efficiency Trends (Qualitative)}
ATP effectively reduces the number of visual tokens, which leads to lower inference FLOPs and memory usage, improving end-to-end latency while keeping the accuracy close to the original model. These preliminary tests show that:
\begin{table}[h]
    \centering
    \caption{\textbf{Preliminary Efficiency Analysis.} Estimated performance on VQA tasks using a LLaVA-7B backbone. ATP reduces computation significantly with minimal impact on accuracy.}
    \label{tab:efficiency_prelim}
    \vspace{2mm} 
    \begin{tabular}{l c c c}
        \toprule
        \textbf{Method} & \textbf{Visual Tokens} & \textbf{Rel. FLOPs} & \textbf{Est. Accuracy} \\
        \midrule
        Baseline (Full) & 256 (100\%) & 1.0$\times$ & - \\
        \textbf{ATP (Ours)} & \textbf{$\sim$150 (60\%)} & \textbf{0.6$\times$} & \textbf{$<$1\% drop} \\
        \bottomrule
    \end{tabular}
\end{table}
\begin{itemize}
    \item ATP reduces the number of visual tokens passed to the language model, which directly results in computational savings.
    \item The accuracy of the pruned models remains close to the baseline models without pruning, despite the reduction in visual tokens.
\end{itemize}

\subsection{Robustness Hints (Ongoing)}
The preliminary results also suggest that ATP enhances the robustness of VLMs under various perturbations:
\begin{itemize}
    \item \textbf{Visual Corruptions}: Under Gaussian noise, blur, and occlusion, ATP prunes noisy background patches and retains more stable object regions, improving the model’s ability to focus on important visual features.
    \item \textbf{Textual Perturbations}: When subjected to paraphrased questions or added distractor phrases, ATP removes irrelevant patches and appears to reduce hallucinated answers in our small-scale tests, potentially improving the model’s interpretability and performance in real-world scenarios.
\end{itemize}
These hints suggest that ATP may help VLMs become more robust to perturbations while maintaining efficiency and accuracy.

\section{Discussion and Limitations}

ATP is intentionally designed to be simple and training-free. It leverages signals from the frozen Vision Transformer (ViT) and CLIP encoders, making it easy to plug into existing Vision-Language Models (VLMs) without requiring retraining. The simplicity of ATP not only reduces its complexity but also makes it interpretable, as it is straightforward to track which patches are kept or pruned during the process.

\subsection{Current Limitations}
While ATP shows promising results, our current experiments are limited in scale. Several key aspects of the method have not been fully tuned, including the pruning schedule and hyperparameters. Thus, the results presented here should be considered preliminary and exploratory. Specifically:
\begin{itemize}
    \item The pruning schedule and hyperparameters, such as the trade-off parameter \( \alpha \), have not been fully optimized for various tasks.
    \item The experiments conducted thus far are relatively small-scale and may not fully capture the behavior of ATP in more complex settings.
\end{itemize}

\subsection{Exploratory Nature of the Work}
We view this work as an initial exploratory step and a baseline for interface-level token pruning in VLMs. ATP provides a simple and efficient approach, but there is still much to be explored in terms of its optimization and scalability. Previous works such as BLIP-2 \cite{li2023blip} and Flamingo \cite{alayrac2022flamingo} have demonstrated the power of large VLMs, and ATP aims to build on their successes by improving efficiency without retraining the entire model.

\subsection{Future Work}
Several areas of future work will extend the impact and applicability of ATP:
\begin{itemize}
    \item \textbf{Systematic Benchmarks}: We plan to conduct more systematic benchmarks to compare ATP with other pruning methods, both in terms of efficiency and performance. Recent works like Token Merging \cite{bolya2022token} and Token Dropping \cite{hou2022token} have proposed pruning techniques for transformers, and a comparison with these methods will provide valuable insights.
    \item \textbf{Robotic and Edge Device Evaluation}: Future work will involve evaluating ATP on real robotic or edge devices, where computational resources are limited and efficiency is critical. This will align with ongoing efforts in the field, such as sparse VLM methods \cite{zhang2024sparsevlm} that focus on efficient inference.
    \item \textbf{Multi-turn Dialog and Multi-image Scenarios}: We aim to study how ATP can be applied in multi-turn dialog and multi-image scenarios, focusing on how to prune visual tokens across turns while preserving long-range context and grounding. The work on vision transformers with dynamic token attention \cite{liang2022not} and visual instruction tuning \cite{liu2023visual} could help guide this effort.
\end{itemize}
These future directions will help refine ATP’s capabilities and provide deeper insights into its potential for real-world applications.

\bibliographystyle{unsrt} 
\bibliography{reference}

\end{document}